\documentclass[10pt,twocolumn,letterpaper]{article}

\usepackage[accsupp]{axessibility}
\usepackage{iccv}
\usepackage{times}
\usepackage{epsfig}
\usepackage{graphicx}
\usepackage{amsmath}
\usepackage{amssymb}

\usepackage{algorithmic}
\usepackage{algorithm}
\usepackage{float}
\usepackage{subfig}
\usepackage{booktabs}
\usepackage{multirow}
\usepackage{pifont}

\usepackage[breaklinks=true,bookmarks=false]{hyperref}

\iccvfinalcopy 

\def\iccvPaperID{10590} 

\ificcvfinal\fi

\begin{document}

\title{Anchor-Intermediate Detector: Decoupling and Coupling Bounding Boxes for Accurate Object Detection}

\author{Yilong Lv, Min Li, Yujie He\\
Xi'an Institute of High Technology\\
{\tt\small \{rfuyll@yeah.net, proflimin@163.com, ksy5201314@163.com\}}
\and
Shaopeng Li\\
Tsinghua University\\
Department of Automation\\
{\tt\small sp-li16@mails.tsinghua.edu.cn}
\and
Zhuzhen He\\
National University of Defense Technology\\
{\tt\small hezhuzhen@163.com}\and
Aitao Yang\\
Xi'an Institute of High Technology\\
{\tt\small  824360083@qq.com}
}

\maketitle
\ificcvfinal\thispagestyle{empty}\fi

\begin{abstract}
	Anchor-based detectors have been continuously developed for object detection. However, the individual anchor box makes it difficult to predict the boundary's offset accurately. Instead of taking each bounding box as a closed individual, we consider using multiple boxes together to get prediction boxes. To this end, this paper proposes the \textbf{Box Decouple-Couple(BDC) strategy} in the inference, which no longer discards the overlapping boxes, but decouples the corner points of these boxes. Then, according to each corner's score, we couple the corner points to select the most accurate corner pairs. To meet the BDC strategy, a simple but novel model is designed named the \textbf{Anchor-Intermediate Detector(AID)}, which contains two head networks, i.e., an anchor-based head and an anchor-free \textbf{Corner-aware head}. The corner-aware head is able to score the corners of each bounding box to facilitate the coupling between corner points. Extensive experiments on MS COCO show that the proposed anchor-intermediate detector respectively outperforms their baseline RetinaNet and GFL method by $\sim$2.4 and $\sim$1.2 AP on the MS COCO test-dev dataset without any bells and whistles. Code is available at: \href{https://github.com/YilongLv/AID}{https://github.com/YilongLv/AID}.
	
\end{abstract}

\section{Introduction}

Object detection is a fundamental and challenging task in computer vision to classify and localize objects in images. Recently, as transformer has achieved good results in machine translation, it has begun to extend into the field of computer vision with success in tasks such as image classification \cite{parmar2018image, liu2021swin, han2021transformer, wang2021pyramid} and object detection \cite{carion2020end, dai2021dynamic, zheng2020end, dosovitskiy2020image}. However, most current mainstream detectors are still based on convolutional neural networks.

\begin{figure}[t]
	\centering
	\subfloat[Raw Image]{
		\begin{minipage}{0.49\linewidth}
			\centering
			\includegraphics[width=1.2in]{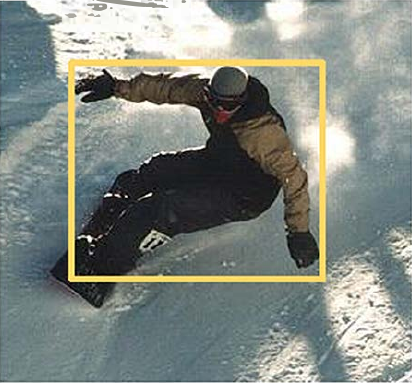}
			\label{fig1-1}
	\end{minipage}}
	\subfloat[Anchor-based]{
	\begin{minipage}{0.49\linewidth}
		\centering
		\includegraphics[width=1.2in]{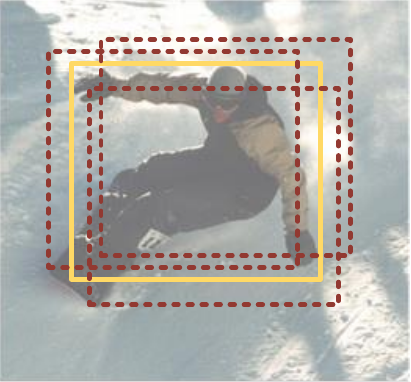}
		\label{fig1-2}
	\end{minipage}}

	\subfloat[Keypoint-base Anchor-free]{
	\begin{minipage}{0.49\linewidth}
		\centering
		\includegraphics[width=1.2in]{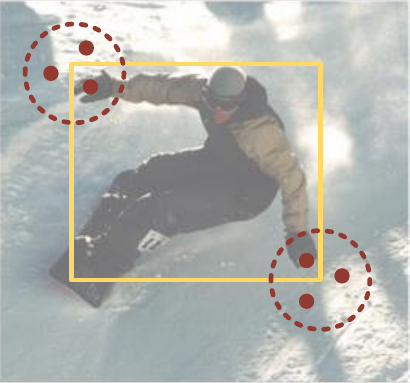}
		\label{fig1-3}
	\end{minipage}}
	\subfloat[Our AID]{
	\begin{minipage}{0.49\linewidth}
		\centering
		\includegraphics[width=1.2in]{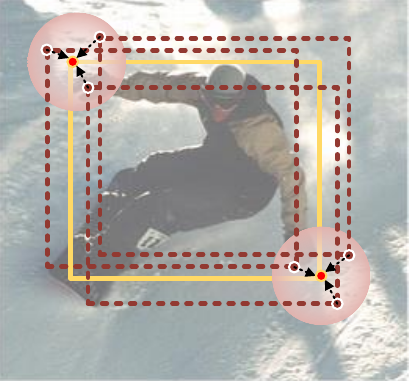}
		\label{fig1-4}
	\end{minipage}}
	\centering
	\caption{The mechanism of the current mainstream detectors. Anchor-based detectors rely on predefined anchor boxes for localization. Anchor-free detectors recombine key points to achieve localization. Our AID takes advantage of both anchor-based and anchor-free methods. It decouples the corner points of the bounding boxes and then, according to the corner's score, the corner points are coupled to select the most accurate corner pair.}
	\label{fig1}
\end{figure}


\begin{figure*}[!t]
	\centering
	\includegraphics[width=5.7in]{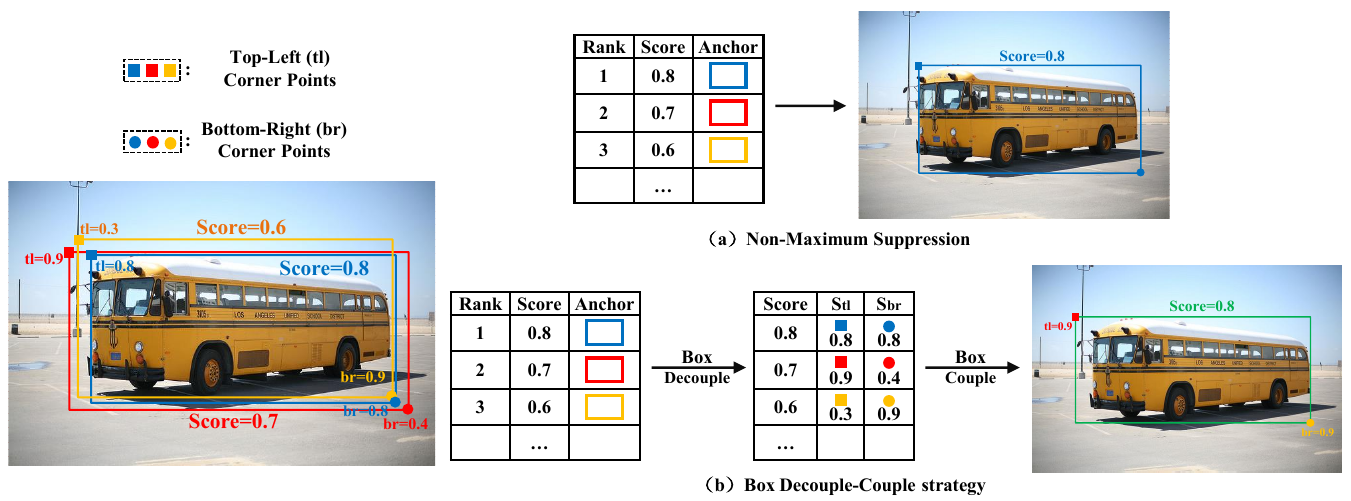}
	\caption{An illustration of the difference between the proposed BDC strategy and existing NMS. For NMS, it retains only the first-ranked predicted box. In contrast, our BDC strategy retains the top-ranked multiple predicted boxes and re-pairs the coordinates according to their corner scores. $tl$ denotes the coordinates of the top-left corner point. $br$ denotes the coordinates of the bottom-right corner point. $S_n$ denotes the score of $n$ corner point. $Score$ denotes the classification score.}
	\label{fig_2}
\end{figure*}


Common anchor-based detectors, such as Faster RCNN \cite{ren2015faster}, Cascade RCNN \cite{cai2018cascade}, YOLO \cite{redmon2016you}, and RetinaNet \cite{lin2017focal}, require pre-defined dense anchor boxes to cover the whole image. Although the anchor box are widely used, it is still a lack of accuracy in locating the object's boundary, because the anchor box is weakly perceptive to the boundary as shown in figure \ref{fig1-2}.

In contrast, some anchor-free method avoid the difficulties of the weak boundary perception.  In particular, the keypoint-based anchor-free detectors such as CornerNet \cite{law2018cornernet}, CenterNet \cite{duan2019centernet}, and CenterNet++ \cite{duan2022centernet++}. Instead of predicting the object's center and boundary' offsets, CornerNet pioneered the corner point prediction mechanism. Concretely, the model decouples each ground truth into left-top and right-down corner points. In this paper, we name this step with a new term, \textbf{Box decouple}. Then, in the inference, CornerNet pairs all the top-left and bottom-right corner points, and this pipeline is shown in figure \ref{fig1-3}, thus forming some bounding boxes. Similarly, we name this step \textbf{Box couple}. Thus the CornerNet don't consider the center-to-boundary perception performance, while improving the localization accuracy. Meanwhile, the net introduces another problem. i.e., the Box couple is very challenging. Suppose the size of the feature map is $\mathbb{R}^{w \times h}$. Then, the number of random corner pairs is $(w \times h)^2$. Too many possible pairs tend to lead to many false positive results, so the average precise drops. CenterNet and CenterNet++ have similar shortcomings. Therefore, how to significantly reduce the number of paired corner points is another challenge we need to address. 

Based on the above analysis, it is found that the keypoint-based anchor-free detectors can circumvent the drawback of anchor-based detectors but also have their attendant difficulties. Therefore, we pondered whether a trade-off can be reached between anchor-based and anchor-free algorithms. Concretely, we can take advantage of the anchor-free detector to improve the shortcomings of the anchor-based detector as shown in figure \ref{fig1-4}. Also, the Box couple dilemma in anchor-free detectors is alleviated by anchor-based detectors.

To this end, this paper proposes a novel architecture named the \textbf{Anchor-Intermediate Detector(AID)}, which is based on the mainstream detection method, including the anchor-based and anchor-free head. First, the anchor-based detection head maintains the conventional training pipeline. Then, we introduce an anchor-free corner-aware head for scoring the corner points of the bounding boxes, making it possible to enhance the boundary perception of the bounding boxes. In detail, during training, the corner-aware head generates two corner-aware heat maps for predicting the distribution of the object's top-left and bottom-right corner points. Similar to CornerNet, but we don't have to predict offsets and classification scores. The AID innovatively integrates two representative detection head frameworks, while the two heads are trained in parallel with each other.

In the inference, we propose a novel post-processing strategy, named \textbf{Box Decouple-Couple (BDC) strategy}. We use the proposed strategy to re-pair the corner points of each prediction box and its overlapping boxes to get more accurate localization results, as shown in figure \ref{fig_2} . In addition, we take into account that the classification score may not be consistent with the localization accuracy, resulting in the prediction results being not most accurately localized. Coincidentally, the corner score can be used as the localization score based on the corner-aware heat map. So, \textbf{Corner-Classification(CoCl) score} are presented, consisting of the classification score and corner confidence for ranking in BDC strategy.

\subsubsection*{\bf The main contributions of our work can be summarized as follows}
\begin{itemize}
	\item{By analyzing the disadvantages of the current anchor-based and anchor-free models, we propose the novel AID, which can achieve a trade-off between the two detection frameworks to improve the detection accuracy of the model.}
	\item{We have redesigned the training and inference pipelines separately. In training, the anchor-based detection head and the corner-aware head share the backbone network and the feature pyramid network. Both are trained synchronously. During inference, we first propose a new corner-classification score for post-processing. Then, our proposed BDC strategy rethinks the value of each prediction box and its overlapping boxes, from which we refine predictions with better localization quality.}
	\item{The AID uses the corner-aware head as the anchor-free head, and the anchor-based head uses some main methods, including RetinaNet, GFL, etc., while based on some backbone networks, including ResNet-50, ResNet-101, ResNeXt-101, etc. Also, our method achieved state-of-the-art results on the MS COCO dataset.}
\end{itemize}

\section{Related Work}
\subsection{Anchor-based detector}
\textbf{Two-stage object detection}. It first started with Faster RCNN, introducing anchor boxes into Fast RCNN \cite{girshick2015fast} using sliding windows and candidate regions. The first stage generates a set of region proposals, and the second stage performs classification and localization fine-tuning on these region proposals. Based on this, many algorithms \cite{he2017mask,dai2016r,cheng2018revisiting, sun2021sparse, he2019bounding} are proposed to improve its performance, feature fusion, attention mechanism, multi-scale training, and training strategy. The two-stage algorithms generally have better detection accuracy but are relatively slow.

\textbf{Single-stage object detection.} For faster detection, single-stage methods have emerged. Instead of relying on RPNs, single-stage detectors directly localized and classified regions of interest in images. SSD \cite{liu2016ssd} was the first approach to propose a single-stage detection strategy, which has attracted much attention due to its efficient training and inference. Since then, many redesigned architectures of single-stage object detectors \cite{redmon2016you,lin2017focal,li2020generalized,zhang2018single,zhang2019freeanchor} have been proposed. Based on them, many methods have been proposed to improve the performance of single-stage detectors, feature fusion, label assignment strategies, detection head network structures, loss functions, and localization refinement. Currently, there are more research results on single-stage than two-stage.

\subsection{Anchor-free detector}
\textbf{Keypoint-based detectors.} The keypoint-based approach focuses on extreme locations of instances, such as corner points and extreme points. CornerNet is one of the representative methods. The improved CornerNet-lite \cite{law2019cornernet} introduces CornerNet-sweep and CornerNet-squeeze to improve its speed. CenterNet adds center points to the top-left and bottom-right corners to provide the ability to perceive internal information, thus improving precision and recall. ExtremeNet \cite{zhou2019bottom} even increases the number of key points to the topmost, bottommost, leftmost, rightmost, and center points. The training pipeline of the anchor-free approach \cite{lu2019grid,yang2019reppoints} is similar to CornerNet.

\textbf{Center-based detector.} These anchor-free methods have similarities with anchor-based methods in that both predict from the center to the boundary. The FCOS \cite{tian2019fcos} method designs a new centrality branching detector head. The centerness scores of every location within the ground truth are also defined. And the label assignment strategy is set accordingly. Different from FCOS, FoveaBox \cite{kong2020foveabox} does not add any new branching network. It regards every position within the subregions of the ground truth as positive and performs label assignment. FSAF \cite{zhu2019feature} connects an anchor-free branch with online feature selection to the RetinaNet. The newly added branch defines the central region of the object as positive and locates it by predicting the four distances to its boundaries. There are also center-based anchor-free detectors \cite{zand2022objectbox, feng2021tood, zhang2020bridging} that play an important role in object detection.

\section{Proposed method}
\begin{figure*}[!t]
	\centering
	\includegraphics[width=6.7in]{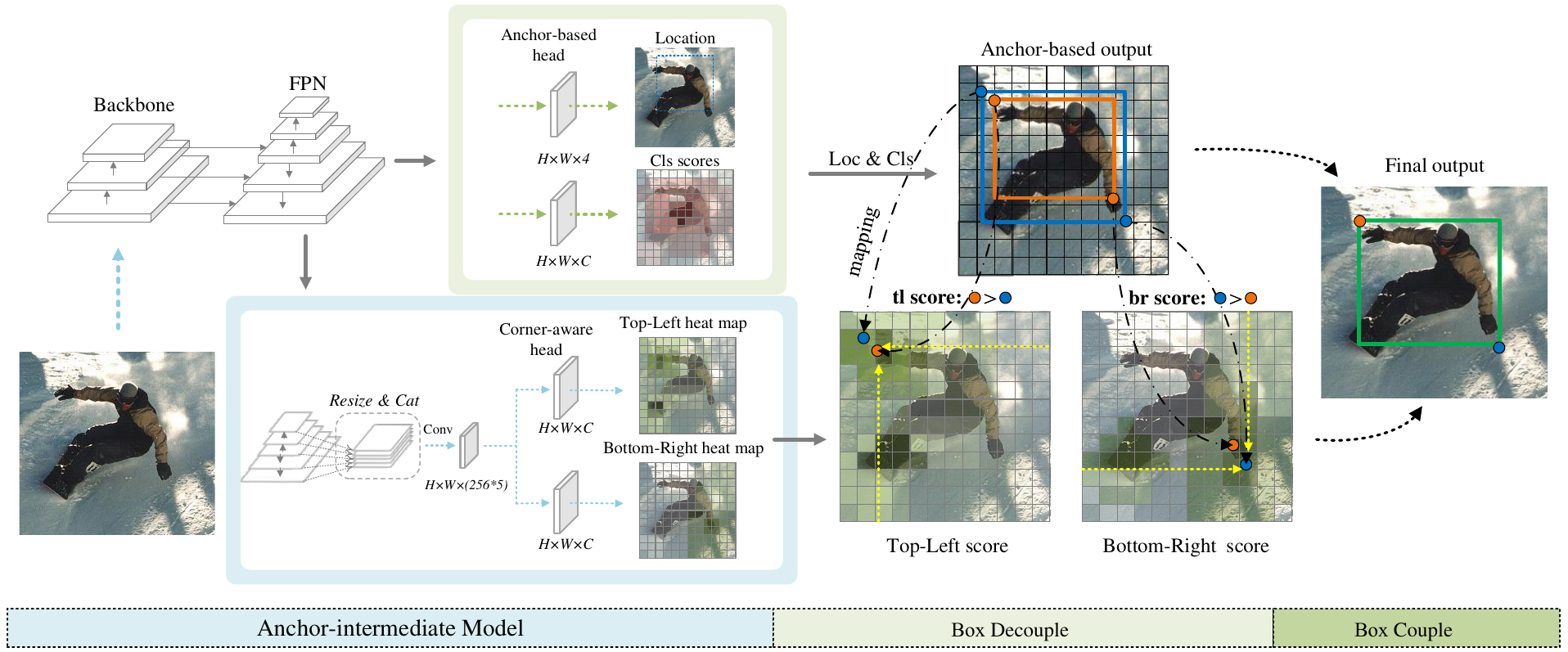}
	\caption{Pipeline of the AID. The left part shows the overall detection model which consists of backbone, FPN and detection head. The upper part is the conventional anchor-based head, which outputs classification scores $(H \times W \times C)$ and boundary prediction $(H \times W \times 4)$, respectively. The lower part is the corner-aware head, which outputs the top-left and bottom-right corner heat maps $(H \times W \times C)$, respectively. In the box decouple, the top-ranked boxes are processed uniformly to separate their corner points. These corner points are scored according to the corner point heat map. Finally, the top-left and bottom-right corner points with good scores are coupled.
	}
	\label{fig_6}
\end{figure*}

In this section, we describe the the AID in detail, including the anchor-based and corner-aware head, as shown in Figure \ref{fig_6}. The anchor-based detection head we use is RetinaNet, but it is not limited to this one. We first introduce the model structure and the training pipeline. Then, in inference, the rules of BDC strategy will be presented in detail.

\subsection{Anchor-Intermediate Detector}
Figure \ref{fig_6} shows our proposed AID, including the corner-aware head and standard anchor-based head. Following the style of multi-task learning, the corner-aware head $f_{cor}$ accepts the features $\mathcal{F}_n$ from the FPN and then learns the corner-aware heat map $\mathcal{M}$. $\mathcal{F}_n$ is the the $n \, (1<=n<=N)$ -th level feature at the FPN. The lower the layer, the smaller the $n$.

Note that we first resize feature so that the multi-scale feature becomes a single scale, as the higher-resolution feature facilitates the differentiation of corner points. Then they were concatenated by channel, which is shown in Eq. \ref{eq1}. Thus, we could obtain single-scale fused features with the architecture shown in Figure \ref{fig_6} of the anchor-intermediate model.
\begin{equation}
	\label{eq1}
	\mathcal{F} = Cat(Resize(\mathcal{F}_n|n=1,\,\dots,\, N)
\end{equation}

\begin{figure}[!h]
	\centering
	\includegraphics[width=3.0in]{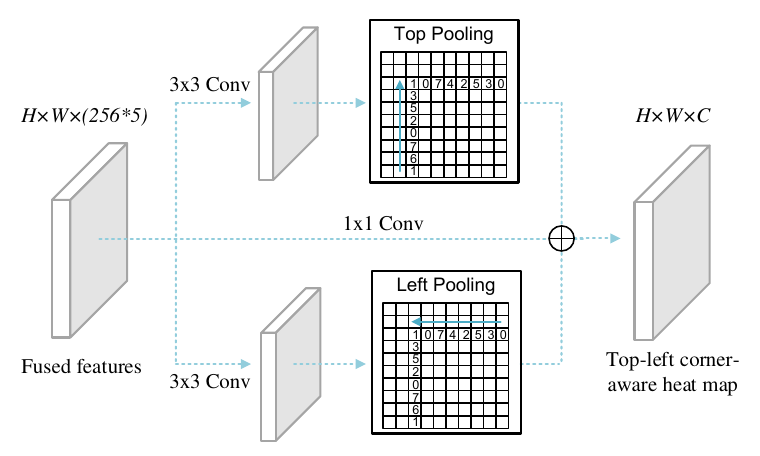}
	\caption{Network structure of the corner-aware head.}
	\label{fig_7}
\end{figure}

For better recognition of corner points, we separately predict the two extreme corners where corner pooling is used. The detection head finally predicts the corner feature heat map $\mathcal{M}=(\mathcal{M}_{tl}, \mathcal{M}_{br})$, including the top-left and bottom-right corner heat maps as $\mathcal{M}^{tl}$ and $\mathcal{M}^{br} \in \mathbb{R}^{w \times h \times c}$. Each set of heatmaps has $C$ channels, which represent the object's categories. 

The structure of the corner-aware head follow the CornerNet, as shown in Figure \ref{fig_7}. The $1\times1$ convolution and activation $\delta$ are performed on the fused features $\mathcal{F}$ according to each direction. Then direction pooling $\mathcal{P} = \{\mathcal{P}_t,\,\mathcal{P}_b,\,\mathcal{P}_r,\,\mathcal{P}_l\}$ is performed separately to obtain the direction-aware feature maps $\{\mathcal{F}_t,\, \mathcal{F}_b,\, \mathcal{F}_r,\, \mathcal{F}_l\}$, and the process is shown in Eq. \ref{eq2}.
\begin{equation}
	\begin{aligned}
		&\{\mathcal{F}_t, \mathcal{F}_l, \mathcal{F}_b, \mathcal{F}_r\} = \mathcal{P}(\delta(Conv(\mathcal{F}))),
	\end{aligned}
	\label{eq2}
\end{equation}

The top-left and bottom-right features are element-wise summed. Then, we use the convolution operation on the features. The feature $\mathcal{F}$ are added to the corner features to prevent a vanishing gradient. Before outputting the heat map, we will perform sigmoid $\sigma$ activation on each element of the feature map so that all positions of the feature map will be between 0 and 1, which represent the confidence. The above is as follows:
\begin{equation}
	\begin{aligned}
		\mathcal{M}_{tl} &= \sigma(Conv(\mathcal{F}_t \oplus  \mathcal{F}_l) + Conv\mathcal{F})\\
		\mathcal{M}_{br} &= \sigma(Conv(\mathcal{F}_b \oplus  \mathcal{F}_r) + Conv\mathcal{F})
	\end{aligned}
\end{equation}

As for the training loss, suppose $\mathcal{M}_{cor}=\{\mathcal{M}_{tl}, \mathcal{M}_{br}\}$ is the corner score in the top-left and bottom-right corner-aware heat map. And let $\mathcal{Y}_{cor}$ be the labeled corner heat map expanded with an unnormalized Gaussian function. The loss function of corner-aware heat map is as follows:
\begin{eqnarray}
	\small
	L_{corner}\!=\!\left\{
	\begin{array}{cl}
		\!\!\!(1-\mathcal{M}_{cor})^{\alpha}log(\mathcal{M}_{cor})      & \!\!\!\! if\, \mathcal{Y}_{cor}=1\\
		\!\!\!\! (1-\mathcal{Y}_{cor})^{\beta}(\mathcal{M}_{cor})^{\alpha}log(1- \mathcal{M}_{cor})       & \!\!\! otherwise
	\end{array} \right.
\end{eqnarray}

Next, the total loss is optimized as shown in Eq. \ref{eq_1}.
\begin{eqnarray}
	\small
	L_{total} = L_{reg} + L_{cls} + \lambda L_{corner}
	\label{eq_1}
\end{eqnarray}
\subsection{CoCl score for post-processing}
In the inference, the overlapping boxes are suppressed based on the classification score $\mathcal{S}_{cls}$ only. However, the classification may not coincide with the localization, which leads to the suppression of the bounding boxes with high localization accuracy when the classification score is low.

We consider corner score as another important basis for post-processing, which is able to integrate the localization information. The corner's score comes from the top-left and bottom-right corner-aware heat map. Therefore, The new evaluation score is named CoCl score, as follows:
\begin{equation}
	\begin{aligned}
		CoCl &= \mathcal{S}_{cls} \times \mathcal{F}(\mathcal{M}_{tl}, \mathcal{M}_{br})\\
	\end{aligned}
	\label{eq5}
\end{equation}

Where $\mathcal{F}$ denotes the integration between $\mathcal{M}_{tl}$ and $\mathcal{M}_{br}$, see section 4.3.3 for details. The results has the same size as $\mathcal{S}_{cls} \in \mathbb{R}^{ijwh \times C}$.
\subsection{Box Decouple-Couple strategy}

\textbf{Box Decouple.} In NMS, the overlapping boxes  are suppressed and then discarded. If a predicted box with highest classification scores fail to locate most accurately, then its overlapping boxes are likely located accurately because they have a large Intersection-over-Union (IoU) with this predicted box. 

The conventional NMS treats each bounding box as an independent individual, but it does not predict the object's location well, especially in those with blurred boundaries. Therefore, we decouple all bounding boxes $\mathcal{B}$ as $\{\mathcal{S}_{tl}, \mathcal{S}_{br}\}$. The details are shown in Eq. \ref{eq6}, where $n$ denotes all predicted bounding boxes, $\mathcal{S}_{tl}$ denotes the top-left point set and $\mathcal{S}_{br}$ denotes the bottom-right point set.
\begin{equation}
	\begin{aligned}
		&\mathcal{B} = (\overbrace{x_{1},\, y_{1}}^{{S}^{tl}},\, \overbrace{x_{2},\, y_{2}}^{{S}^{br}}),\\
		&\mathcal{S}_{tl} =\{s^{tl}_i\}_{i=1, \dots ,n},\,\,\,\,\,\mathcal{S}_{br}=\{s^{br}_i\}_{i=1, \dots ,n}.
	\end{aligned}
	\label{eq6}
\end{equation}

For the predicted bounding box, suppose the box $\mathcal{P}$ corresponds to the overlapping box  $\mathcal{O}=\{b \,| \, IoU(b,\mathcal{P})>\tau ,\, b \in \mathcal{B}\}$. We decouple the predicted box $\mathcal{P}$ and its overlapping boxes $\mathcal{O}$ into the top-left corner set $\{s^{tl}_p,s^{tl}_{o1}, \dots, s^{tl}_{oi}\} \subseteq \mathcal{S}_{tl}$ and bottom-right corner sets $\{s^{br}_p, s^{br}_{o1}, \dots, s^{br}_{oi}\} \subseteq \mathcal{S}_{br}$. Then, we map these corner points onto the corner-aware heat map $\mathcal{M}$ and the process is as follows:
\begin{equation}
	\begin{aligned}
		&(f^{br}_p, f^{br}_{o1}, \dots, f^{br}_{oi}) = f_{\mathcal{P,O} \mapsto \mathcal{M}_{br}}(s^{br}_p, s^{br}_{o1}, \dots, s^{br}_{oi})\\
		&(f^{tl}_p, f^{tl}_{o1}, \dots, f^{tl}_{oi}) = f_{\mathcal{P,O} \mapsto \mathcal{M}_{tl}}(s^{tl}_p, s^{tl}_{o1}, \dots, s^{tl}_{oi})
	\end{aligned}
\end{equation}

Therefore, we use box decoupling to transition the anchor-based inference to anchor-free inference, as shown in Figure \ref{fig_6} of box decouple. Here, these decoupled corner points are very similar to those in CornerNet. Next, we use the idea of anchor-free to process these corner points. It is worth emphasizing that our method dramatically reduces the number of corner point pair in our method, thus improving the average precise. 

\textbf{Box Couple.} Since the heat maps $\mathcal{M}$ are downsampled, each element  in the corner-aware heat map does not correspond to the original image one by one, but is mapped to a region of image. In addition, the corner-aware heat map is fitted with a Gaussian model in the training, so the confidence for pixels at the same distance from the center is the same. Therefore, there are some locations in the heat map with the same confidence. To this end, we chose multiple decoupled corner points and use them to obtain a new prediction box. This process is specified in Eq. \ref{eq8}.
\begin{equation}
	\begin{aligned}
		\mathcal{T}^{br} = &\underset {s^{br}_p, s^{br}_{o1}, \dots, s^{br}_{oi}}{\arg\max}\,\mathop{topk}^{n}(f^{br}_p, f^{br}_{o1}, \dots, f^{br}_{oi}) \\
		\mathcal{T}^{tl} = &\underset {s^{tl}_p, s^{tl}_{o1}, \dots, s^{tl}_{oi}}{\arg\max}\, \mathop{topk}^{n}(f^{tl}_p, f^{tl}_{o1}, \dots, f^{tl}_{oi})
	\end{aligned}
	\label{eq8}
\end{equation}

Finally, the new corner points obtained are combined to form an updated bounding box $\mathcal{B}_{update}$. The details are shown in Eq. \ref{eq9}. 
\begin{equation}
	\begin{aligned}
		\mathcal{B}_{update} &= (Mean(\{s^{tl}_i\}_{i=\mathcal{T}^{br}}), \,Mean(	\{s^{tl}_i\}_{i=\mathcal{T}^{tl}}))
	\end{aligned}
	\label{eq9}
\end{equation}

Following this method, the output boxes not only contain classification and localization's information but also improve the accuracy of localization. The detailed procedure of the BDC strategy is shown in Algorithm \ref{alg:alg1}.

\begin{algorithm}[H]
	\caption{Box Decouple-Couple strategy.}\label{alg:alg1}
	\begin{algorithmic}
		\STATE
		\STATE {\textsc{TRAIN:}} Corner heat map $(\mathcal{M}_{tl}, \mathcal{M}_{br})$
		\STATE \hspace{0.5cm} $\textbf{select } \mathcal{S}_{ijc} \textbf{ for }  i,j,c, \mathcal{M}$ \textbf{in} $W,H,C, [\mathcal{M}_{tl},\mathcal{M}_{br}]$
		\STATE \hspace{0.5cm} \textbf{ if } $\mathcal{Y}_{ijc}==1$\textbf{: }
		\STATE \hspace{1cm} $(1-\mathcal{S}_{ijc})^{\alpha}log(\mathcal{S}_{ijc})\mapsto 0$
		\STATE \hspace{0.5cm} \textbf{ else: }
		\STATE \hspace{1cm} $(\mathcal{S}_{ijc})^{\alpha}log(1- \mathcal{S}_{ijc})\mapsto 0$
		\STATE 
		\STATE {\textsc{PREDICT:}}
		\STATE \hspace{0.5cm} $CoCl = \mathcal{S}_{cls} \times \mathcal{F}(\mathcal{M}_{tl}, \mathcal{M}_{br})$
		\STATE \hspace{0.5cm} $\mathcal{B}^p, \mathcal{B}^o$ \textbf{ = NMS($CoCl$)} 
		\STATE \hspace{0.5cm} \textbf{\# $\mathcal{B}^p \in \mathbb{R}^{1\times4}$ $\Rightarrow$Prediction Box}
		\STATE \hspace{0.5cm} \textbf{\# $\mathcal{B}^o \in \mathbb{R}^{N\times4}$ $\Rightarrow$ Overlapping  Box}
		\STATE \hspace{0.5cm} $\{\mathcal{S}^{tl}, \mathcal{S}^{br}\}$ = \textbf{Decouple}($\mathcal{B}^p_{n},\mathcal{B}^o_{n}$)
		\STATE \hspace{0.5cm} 	$(f^{br}, f^{tl}) = f_{\mathcal{P,O} \mapsto \mathcal{M}}(\mathcal{S}^{tl}, \mathcal{S}^{br})$
		\STATE \hspace{0.5cm} $\textbf{select the top-n } [f^{br}, f^{tl}] \mapsto [f^{br}_{top-n}, f^{tl}_{top-n}]$ 
		\STATE \hspace{0.5cm} 	$(\mathcal{S}^{tl}_{top-n}, \mathcal{S}^{br}_{top-n}) = f_{ \mathcal{M} \mapsto \mathcal{P,O}}(f^{br}_{top-n}, f^{tl}_{top-n})$
		\STATE \hspace{0.5cm} $ \mathcal{B}_{update} = \textbf{Couple}(\mathcal{S}^{tl}_{top-n}, \mathcal{S}^{br}_{top-n}) $
		\STATE \hspace{0.5cm}\textbf{return}  $ \mathcal{B}_{update} $
	\end{algorithmic}
	\label{alg1}
\end{algorithm}

\section{Experiment}
\subsection{Dataset}
We evaluate our method on the MS-COCO dataset \cite{lin2014microsoft} according to the commonly used settings. MS-COCO contains about 160K images of 80 classes. The dataset was partitioned into training 2017, val2017, and test 2017 subsets with 118K, 5K, and 41K images, respectively. The standard average precision (AP) metric reports the results at different IoU thresholds and target scales. We trained only on the 2017 train images in all our experiments without using any additional data. For the experiments of the ablation study, we evaluated the performance on a subset of val2017. Compared to state-of-the-art methods, we report the official results returned from the test server on the test-dev subset.

\subsection{Implementation details}
We use the generic "Backbone - FPN - Head" as our pipeline and the MMdetection toolbox \cite{chen2019mmdetection} to implement our method. All models are trained on 4 TESLA A100 GPUs with four small batches per GPU. Unless otherwise specified, we used a stochastic gradient descent (SGD) optimizer with a weight decay of 0.0001 and momentum of 0.9. The initial learning rate was set to 0.01, and training was started using a linear warm-up strategy. We initialize our backbone network with the weights pre-trained on ImageNet \cite{deng2009imagenet}.

\subsection{Ablation study}
This section will perform detailed ablation experiments on the proposed method. Besides exploring the effects of different baselines and backbone networks on the experimental results, the rest of the experiments are performed on the RetinaNet method based on ResNet-101 and trained for 12 epochs. The final validation is performed on the COCO-val2017 dataset.

\subsubsection{Corner-aware head}
In this section, we discuss the results after adding corner-aware head at different positions of the model, and the experimental results are shown in Table \ref{tab1}. From the table, the performance of our method improves by 0.9\% AP on the classification branch and by 0.6\% AP on the regression branch.

\begin{table}[h]
	\centering
	\small
	\setlength{\tabcolsep}{1.0mm}{
		\begin{tabular}{|l|cccccc|}
			\hline
			Method & AP   & AP50 & AP75 & APS  & APM  & APL  \\ \hline
			\hline
			Baseline                  & 38.5 & 57.6 & 41   & 21.7 & 42.8 & 50.4 \\
			+ Cor head - FPN                    & 39.8(+1.3) & 58.7 & 42.4 & 21.8 & 43.5 & 52.8 \\
			+ Cor head - cls                    & 39.4(+0.9) & 58.4 & 42.4 & 21.4 & 43.5 & 53.2 \\
			+ Cor head - reg                    & 39.1(+0.6) & 58.6 & 42.1 & 21.4 & 42.9 & 52.6 \\ \hline
	\end{tabular}}
	\caption{Comparison of performances when applying the our method to each position of the baseline model.\label{tab1}}
\end{table}

We perform the analysis. Corner-aware head belongs to the anchor-free head. In contrast, the classification and regression branches belong to the anchor-based head, and it seems more common sense for them to be trained independently. Therefore, we also separate the Corner-aware head network from the two branches and connect it to the back of the FPN. This way, the Corner-aware head and Cls-Reg branches will present a parallel structure. The performance of this structure is improved by 1.3\% AP. The experimental results illustrate that separating the Corner-aware head from the classification and regression branches can improve the model's performance by making them independent.

\subsubsection{Hyperparameters $\lambda$}
Further, we discuss the effect of $\lambda$ in Eq. \ref{eq_1} on the experimental results, shown in Table \ref{tab3}. When the hyperparameters $\lambda$ is equal to 0, it indicates a baseline method without the corner-aware head, which has a detection performance of 38.5 AP. The detection performance of the AID reaches the highest 40.1 AP when the hyperparameters $\lambda$ is equal to 0.3. Meanwhile. The AP decreases as the $\lambda$ keeps increasing, which indicates that if the corner-aware head is trained with large weights, it will adversely affect the training of the traditional detection training loss, thus leading to a decrease in detection performance.

\begin{table}[h]
	\centering
	\small
	\setlength{\tabcolsep}{0.75mm}{
		\begin{tabular}{|l|cccccc|}
			\hline
			Weight factor & AP   & AP50 & AP75 & APS  & APM  & APL  \\ \hline
			\hline 
			0(baseline)                             & 38.5 & 57.6 & 41   & 21.7 & 42.8 & 50.4 \\
			0.1                           & 39.8(+1.3) & 58.7 & 42.4 & 21.8 & 43.5 & 52.8 \\
			0.3                           & 40.1(+1.6) & 58.8 & 42.8 & 21.8 & 43.6 & 53.2 \\
			0.5                           & 39.4(+0.9) & 58.1 & 42.2 & 22.2 & 42.5 & 52.2 \\
			0.8                           & 39.0(+0.5) & 56.8 & 40.7 & 20.5 & 41.4 & 50.7 \\
			1                             & 38.2(-0.3) & 56.8 & 40.7 & 20.5 & 41.4 & 50.7 \\ \hline
	\end{tabular}}
	\caption{Peformance of the AID when changing the
		hyperparameters $\lambda$ of the total loss. $\lambda$ weighting
		means weighting the loss of the corner-aware head.\label{tab3}}
\end{table}

\subsubsection{CoCl score}
During the inference, the detection confidence is calculated according to the classification score. We next discuss the contribution of several classical forms of $\mathcal{F}(\mathcal{M}_{tl}, \mathcal{M}_{br})$ to the detection accuracy, and the experimental results are shown in Table \ref{tab4}. First, the detection result obtained by multiplying the classification score and the corner score is 39.6, when the corner score take the maximum output of the top-left and bottom-right corner-aware heat map. Also, calculating the minimum output of the two corner-aware heat map as the corner score is a scheme with a detection accuracy of 39.6 AP. Finally, the detection performance AP is 39.8 when the average output of both heat map is calculated.

\begin{table*}[t]
	\centering
	\small
	\setlength{\tabcolsep}{2.0mm}{
		\begin{tabular}{|c|c|cc|cc|cccccc|}
			\hline
			\multirow{2}{*}{Method} &
			\multirow{2}{*}{Backbone} &
			\multicolumn{2}{c|}{AID} &
			\multirow{2}{*}{PAFPN} &
			\multirow{2}{*}{ATSSAssigner} &
			\multirow{2}{*}{AP} &
			\multirow{2}{*}{AP50} &
			\multirow{2}{*}{AP75} &
			\multirow{2}{*}{APS} &
			\multirow{2}{*}{APM} &
			\multirow{2}{*}{APL} \\ \cline{3-4}
			&
			&
			CoCl score &
			BDC &
			&
			&
			&
			&
			&
			&
			&
			\\ \hline \hline
			\multicolumn{1}{|l|}{Retinanet} &
			\multirow{7}{*}{ResNet-101} &
			&
			&
			&
			&
			38.5 &
			57.6 &
			41 &
			21.7 &
			42.8 &
			50.4 \\ \cline{1-1} \cline{3-12} 
			\multicolumn{1}{|c|}{\multirow{6}{*}{AID}} &
			&
			\ding{52} &
			&
			&
			&
			\multicolumn{1}{c}{39.8(+1.3)} &
			\multicolumn{1}{c}{58.7} &
			\multicolumn{1}{c}{41.1} &
			\multicolumn{1}{c}{20.6} &
			\multicolumn{1}{c}{42} &
			51.2 \\
			\multicolumn{1}{|c|}{} &
			&
			&
			\ding{52} &
			&
			&
			\multicolumn{1}{l}{39.1(+0.6)} &
			\multicolumn{1}{l}{58.3} &
			\multicolumn{1}{l}{41.5} &
			\multicolumn{1}{l}{21.6} &
			\multicolumn{1}{l}{43.1} &
			\multicolumn{1}{l|}{51.8} \\
			\multicolumn{1}{|c|}{} &
			&
			\ding{52} &
			\ding{52} &
			&
			&
			40.0(+1.5) &
			58.7 &
			42.1 &
			31.6&
			43.3&
			52.8 \\ \cline{3-12} 
			\multicolumn{1}{|c|}{} &
			&
			\ding{52} &
			\ding{52} &
			\ding{52} &
			&
			40.1(+1.6) &
			58.8 &
			43 &
			22.1 &
			43.9 &
			53.1 \\
			\multicolumn{1}{|c|}{} &
			&
			\ding{52} &
			\ding{52} &
			&
			\ding{52} &
			40.6(+2.1) &
			57.3 &
			43.9 &
			22.9 &
			44.4 &
			52.3 \\
			\multicolumn{1}{|c|}{} &
			&
			\ding{52} &
			\ding{52} &
			\ding{52} &
			\ding{52} &
			\multicolumn{1}{l}{40.8(+2.3)} &
			\multicolumn{1}{l}{57.4} &
			\multicolumn{1}{l}{44.3} &
			\multicolumn{1}{l}{23.5} &
			\multicolumn{1}{l}{44.4} &
			\multicolumn{1}{l|}{52.4} \\ \hline
	\end{tabular}}
	\caption{Individual contribution of the components in our method. The first row represents the results of the baseline trained with the focal loss. PAFPN is used as a more powerful feature pyramid network. ATSSAssigner is used to replace the conventional MaxIoUAssigner.\label{tab8}}
\end{table*}

\begin{table}[h]
	\centering
	\small
	\setlength{\tabcolsep}{1.2mm}{
		\begin{tabular}{|l|cccc|}
			\hline
			id & F & AP   & AP50 & AP75   \\ \hline \hline
			1  & $\mathcal{S} \times e^{avg(\mathcal{M}_{tl},\mathcal{M}_{br})}$ & 39.8 & 58.7 & 42.4  \\
			3  & $\mathcal{S} \times e^{max(\mathcal{M}_{tl},\mathcal{M}_{br})}$ & 39.6 & 58.6 & 42.2  \\
			3  & $\mathcal{S} \times e^{min(\mathcal{M}_{tl},\mathcal{M}_{br})}$ & 39.6 & 58.5 & 42.2  \\
			4  & $\mathcal{S}^0 \times avg(\mathcal{M}_{tl},\mathcal{M}_{br})^{1}$ & 39.0 & 58.3 & 41.5  \\
			5  & $\mathcal{S}^{0.3} \times avg(\mathcal{M}_{tl},\mathcal{M}_{br})^{0.7}$ & 39.9 & 58.5 & 42.7  \\
			6  & $\mathcal{S}^{0.5} \times avg(\mathcal{M}_{tl},\mathcal{M}_{br})^{0.5}$ & 39.1 & 57.1 & 41.8 \\ 7 & $\mathcal{S}^{0.8} \times avg(\mathcal{M}_{tl},\mathcal{M}_{br})^{0.2}$ & 35.7 & 52.0 & 38.2  \\
			8 & $\mathcal{S}^{1} \times avg(\mathcal{M}_{tl},\mathcal{M}_{br})^{0}$ & 28.5 & 44.3 & 29.2  \\ \hline
	\end{tabular}}
	\caption{Comparison of performances of different CoCl score functions.\label{tab4}}
\end{table}

In addition, We also perform a weighted average of the classification score and the corner score, using a balance parameter $\alpha$ to adjust the ratio between them, as shown in Eq. \ref{eq11}. When $\alpha$ equals to 1, the detection confidence degenerates to the classification score. Similarly, the detection confidence degenerates to the corner score when we $\alpha$ equals to 0. In the inference, we studied the different parameters. The experimental results are shown in Table \ref{tab4}. From the experimental results, the method in this paper can achieve 0.3 when $\alpha$ is equal to 39.9.

\begin{equation}
	\begin{aligned}
		CoCl &=\mathcal{S}_{cls}^{\alpha} \times avg(\mathcal{M}_{tl},\mathcal{M}_{br})^{\alpha}
	\end{aligned}
	\label{eq11}
\end{equation}


\subsubsection{Box Decouple-Couple strategy}
In the inference, for the prediction boxes and the overlapping boxes, we next investigate the contribution of the strategies of different box coupling to the experimental results, which are shown in Table \ref{tab5}. The simple strategy is to select the highest detection score, which has a detection accuracy of 39.7 AP, 1.2 AP higher than the baseline method. We continued our study by averaging all the top-left and bottom-right corner points. Its detection result is 36.0 AP, which is lower than the performance of the baseline method. We analyze that because some corner points with small detection scores are not suitable for predicting object. Therefore, we consider the sum of the mean and deviation of the detection scores as the threshold value $\tau_{score}$. The positions of the corner points with detection scores larger than $\tau_{score}$ are averaged, and the detection performance under this strategy is 40.0 AP, which is 1.5 AP higher than the baseline method. Also, we average the corner points with the top n(=10) scores, resulting in 39.8 AP, which is 1.3 AP higher than the baseline method.

\begin{table}[h]
	\centering
	\small
	\setlength{\tabcolsep}{1.0mm}{
		\begin{tabular}{|l|ccccccc|}
			\hline
			id & F & AP   & AP50 & AP75 & APS  & APM  & APL  \\ \hline \hline
			1  & Top-n(=10) & 39.8(+1.3) & 58.6 & 41.1   & 20.6 & 42 & 51.2 \\
			2  & Max & 39.7(+1.2) & 58.7 & 42.1 & 21.6 & 43.3 & 52.8 \\
			3  & All & 36.0(-2.5) & 58.6 & 39 & 20.6 & 40.6 & 46.2 \\
			4  & $\tau_{score}(=0.5)$ & 40.0(+1.5) & 58.7 & 42.7 & 21.7 & 43.7 & 53.4 \\ \hline
	\end{tabular}}
	\caption{Comparison of performances of different box couple method. \textbf{Max} means to select the corner point with the maximum score. \textbf{Top-n} means select the first n maximum corner point. \textbf{All} means select all corner points. \textbf{$\tau_{score}(=0.5)$} means to select the corner point coordinates greater than the threshold (=0.5).\label{tab5}}
\end{table}

\subsubsection{Stronger components}



We conducted experiments using the components of the BDC strategy, different feature pyramid networks and label assignment strategies to improve the detection performance of the AID further, which are shown in Table \ref{tab8}. When PAFPN is used as the feature pyramid network, our method achieves 40.1 AP, which improves 1.6 AP compared to the baseline method. In addition, we use ATSSAssigner alone instead of the conventional maximum IoU strategy. Our method improves 40.6 AP compared to the baseline method. When the PAFPN and ATSSAssigner are used together, the proposed method achieves 40.8 AP.

%

\subsection{Comparison with State-of-the-Arts}
For experiments comparing with the state-of-the-art dense detector on COCO test-dev2017, we train 1x and 2x (i.e., 12 and 24 epochs) models.

\begin{table*}[htbp]
	\centering
	\small
	\setlength{\tabcolsep}{2.8mm}{
		\begin{tabular}{|l|cccccccc|}
			\hline
			Method                      & Backbone           & Schedule & AP   & AP50 & AP75 & APS  & APM  & APL   \\ \hline \hline
			Anchor-based two-stage:     &                    &          &      &      &      &      &      &       \\
			Faster RCNN \cite{ren2015faster}                 & ResNet-101         & 2x       & 39.7 & 60.7 & 43.2 & 22.5 & 42.9 & 49.9  \\
			Cascade R-CNN \cite{cai2018cascade}            & ResNet-101         & 2x       & 42.8 & 62.1 & 46.3 & 23.7 & 45.5 & 55.2  \\
			Mask R-CNN \cite{he2017mask}                  & ResNet-101         & 2x       & 39.8 & 62.3 & 43.4 & 22.1 & 43.2 & 51.2  \\
			R-FCN   \cite{dai2016r}                    & ResNet-101         & 2x       & 41.4 & 63.4 & 45.2 & 24.5 & 44.9 & 51.8  \\
			TridentNet  \cite{paz2022tridentnet}                & ResNet-101         & 2x       & 42.7 & 63.6 & 46.5 & 23.9 & 46.6 & 56.6  \\ \hline \hline
			Anchor-free Keypoint-based: &                    &          &      &      &      &      &      &       \\
			CornerNet  \cite{law2019cornernet}            & hourglass-104      & 200e       & 40.5 & 56.6 & 43.1 & 18.9 & 43   & 53.4  \\
			CenterNet \cite{duan2019centernet}                  & hourglass-104      & 190e       & 44.9 & 62.4 & 48.1 & 25.6 & 47.4 & 57.4  \\
			ExtremeNet    \cite{zhou2019bottom}              & hourglass-104      & 200e       & 40.2 & 55.5 & 43.2 & 20.4 & 43.2 & 53.1  \\
			Grid R-CNN \cite{lu2019grid}                 & ResNet-101         & 2x       & 43.2 & 63.0 & 46.6 & 25.1 & 46.5 & 55.2  \\
			RepPoints \cite{yang2019reppoints}                 & ResNet-101-dcn     & 2x       & 45.0 & 66.1 & 49.0 & 26.6 & 48.6 & 57.5  \\ \hline \hline
			Anchor-free Center-based:   &                    &          &      &      &      &      &      &       \\
			FCOS      \cite{tian2019fcos}                  & ResNet-101         & 2x       & 43.0 & 61.7 & 46.3 & 26.0 & 46.8 & 55.0  \\
			FoveaBox    \cite{kong2020foveabox}                & ResNeXt-101        & 2x       & 42.1 & 61.9 & 45.2 & 24.9 & 46.8 & 55.6  \\
			FSAF \cite{zhu2019feature}                       & ResNext-101-64x4d & 2x       & 42.9 & 63.8 & 46.3 & 26.6 & 46.2 & 52.7  \\ \hline \hline
			Anchor-based one-stage:     &                    &          &      &      &      &      &      &       \\
			SSD512 \cite{liu2016ssd}                     & VGG16              & 2x       & 28.8 & 48.5 & 30.3 & 10.9 & 31.8 & 43.5  \\
			RefineDet \cite{zhang2018single}                  & ResNet-101         & 2x       & 41.8 & 62.9 & 45.7 & 25.6 & 45.1 & 54.1  \\
			FreeAnchor \cite{zhang2019freeanchor}                 & ResNet-101         & 2x       & 43.1 & 62.2 & 46.4 & 24.5 & 46.1 & 54.8  \\ \hline \hline
			Baseline:                   &                    &          &      &      &      &      &      &       \\
			RetinaNet  \cite{lin2017focal}                 & ResNet-50          & 1x       & 36.9 & 56.2 & 39.3 & 20.5 & 39.9 & 46.3  \\
			RetinaNet                   & ResNet-50          & 2x       & 37.7 & 57.2 & 40.2 & 20.4 & 40.2 & 48.1  \\
			RetinaNet                   & ResNet-101         & 1x       & 39   & 58.6 & 41.7 & 21.9 & 42.2 & 49.3  \\
			RetinaNet                   & ResNet-101         & 2x       & 39.6 & 59.1 & 42.4 & 21.3 & 42.6 & 51.1  \\
			RetinaNet                   & ResNeXt-101-32x4d & 1x       & 40.4 & 60.3 & 43.1 & 23   & 43.7 & 50.6  \\
			RetinaNet                   & ResNeXt-101-64x4d & 1x       & 41.4 & 61.5 & 44.4 & 24.2 & 44.8 & 52.1  \\
			GFL        \cite{li2020generalized}                 & ResNet-50          & 1x       & 40.5 & 58.9 & 43.8 & 22.8 & 43.6 & 50.6  \\
			GFL                         & ResNet-50          & 2x       & 43.2 & 61.7 & 46.9 & 26.5 & 46.7 & 51.8  \\
			GFL                         & ResNet-101         & 2x       & 45.2 & 63.9 & 49.3 & 27.5 & 48.8 & 55.3  \\
			GFL                         & ResNet-101-dcn     & 2x       & 47.3 & 66   & 51.5 & 28.3 & 50.9 & 59.3  \\
			GFL                         & ResNeXt-101-dcn     & 2x       & 48.2 & 67.2 & 52.5 & 29.2 & 51.6 & 60.2  \\ \hline \hline
			Ours:                       &                    &          &      &      &      &      &      &       \\
			RetinaNet+AID                  & ResNet-50          & 1x       & 38(+1.1)   & 56.9 & 40.6 & 20.8 & 40.4 & 47.8  \\
			RetinaNet+AID                 & ResNet-50          & 2x       & 39.2(+1.5) & 58.3 & 41.8 & 21.1 & 41.4 & 49.9  \\
			RetinaNet+AID                 & ResNet-101         & 1x       & 40.1(+1.1) & 59.4 & 42.9 & 22.2 & 42.9 & 51    \\
			RetinaNet+AID                   & ResNet-101         & 2x       & 41.2(+1.6) & 60.5 & 44   & 22.3 & 43.8 & 53    \\
			RetinaNet+AID                   & ResNeXt-101-32x4d & 1x       & 42.8(+2.4) & 62.6 & 46   & 24.6 & 45.8 & 54.3  \\
			RetinaNet+AID                  & ResNeXt-101-64x4d & 1x       & 41.7(+0.3) & 61.3 & 44.6 & 23.6 & 44.4 & 52.88 \\
			GFL+AID                         & ResNet-50          & 1x       & 41.1(+0.6) & 58.9 & 44.3 & 23.2 & 43.9 & 51    \\
			GFL+AID                        & ResNet-50          & 2x       & 43.9(+0.7) & 62.1 & 47.7 & 26.7 & 47.1 & 52.8  \\
			GFL+AID                         & ResNet-101         & 2x       & 45.7(+0.5) & 64.1 & 49.7 & 27.6 & 49.3 & 55.7  \\
			GFL+AID                         & ResNet-101-dcn     & 2x       & 48.4(+1.1) & 66.9 & 52.6 & 29   & 51.9 & 60.8  \\
			GFL+AID                         & ResNeXt-101-dcn     & 2x       & 49.4(+1.2) & 67.9 & 53.6 & 29.6 & 53   & 62    \\ \hline
	\end{tabular}}
	\caption{Performance comparison with the state-of-the-art methods on the MS-COCO
		test-dev dataset in single-model and single-scale results. Our method respectively outperforms their baseline RetinaNet and GFL method by $\sim$2.4 and $\sim$1.2 AP without any bells and whistles. \label{tab9}}
\end{table*}

To demonstrate the effectiveness of the proposed method, we have conducted a series of experiments based on two baselines and various advanced backbone networks. The results are presented in Table \ref{tab9}, from which it can be seen that our model achieves excellent performance. As shown in Table \ref{tab9}, using ResNet-101 and ResNeXt-101-64$\times$4d, our method based on the RetinaNet model achieves 40.1$\sim$41.7 AP, consistently outperforming current baseline. Meanwhile, using ResNet-101 and ResNeXt-101-64$\times$4d, our method based on the GFL model achieves 45.7$\sim$49.4 AP, which outperforms the baseline GFL methods by approximately 0.3$\sim$1.2 AP.

\section{Conclusion}
We focus on improving the performance of the anchor-based detector. The novel AID is proposed and contains three innovations: Firstly, The corner-aware head is proposed to quantify the localization of each corner point. Then, The BDC strategy is proposed. The overlapping boxes are fully utilized to decouple the corner points, and then the corner points are re-paired. In addition, the CoCl score is proposed, which contains both classification and corner scores and can comprehensively evaluate the quality of prediction boxes. Experimental results demonstrate the effectiveness of these improvements. The performance of our method exceeds that of many state-of-the-art models. However, there are improvements in our work: The corner-aware head increases the training burden of the model, and we will consider using knowledge distillation to achieve a lighter model. And, the BDC strategy has many manually predefined hyperparameters, and adaptive hyperparameters will be designed in the future.

{\small
\bibliographystyle{ieee_fullname}
\bibliography{egbib}
}

\end{document}